\documentclass[lettersize,journal]{IEEEtran}
\usepackage{amsmath,amsfonts}
\usepackage{algorithmic}
\usepackage{algorithm}
\usepackage{array}
\usepackage[caption=false,font=normalsize,labelfont=sf,textfont=sf]{subfig}
\usepackage{textcomp}
\usepackage{stfloats}
\usepackage{url}
\usepackage{verbatim}
\usepackage{graphicx}
\usepackage{cite}
\usepackage[pagebackref,breaklinks,colorlinks]{hyperref}
\usepackage{annotate-equations}
\usepackage{subcaption}
\usepackage{pifont}%
\usepackage{placeins}
\usepackage{tabularx}
\usepackage{booktabs} %
\usepackage{soul}
\usepackage{multirow}
\usepackage{colortbl}
\usepackage[inkscapelatex=false]{svg}

\definecolor{myorange}{HTML}{FDE8D4}
\definecolor{myblue}{HTML}{D4D4FF}
\definecolor{mygreen}{HTML}{D4FFD4}
\definecolor{mygred}{HTML}{F8D5D5}
\newcommand{\rv}[1]{#1}

\def\eg{\emph{e.g}, } 

\def\ie{\emph{i.e, }}

\belowdisplayskip \abovedisplayskip
\begin{document}

\title{InstructHumans: Editing Animated\\ 3D Human Textures with Instructions}

\author{%
Jiayin Zhu, Linlin Yang, Angela Yao,~\IEEEmembership{Member,~IEEE}
%
\thanks{\noindent Jiayin Zhu and Angela Yao, National University of Singapore, 117418, Singapore. (email: zhujiayin@u.nus.edu; ayao@comp.nus.edu.sg).}
\thanks{\noindent Linlin Yang, Communication University of China, Beijing, 100024, China. (email: lyang@cuc.edu.cn).}
}

\markboth{IEEE TRANSACTIONS ON MULTIMEDIA,~Vol.~XX, No.~X, ~2025}%
{Jiayin Zhu, Linlin Yang, \MakeLowercase{\textit{et al.}}: InstructHumans: Editing Animated  3D Human Textures with Instructions}


\maketitle

\begin{abstract}
We present InstructHumans, a novel framework for instruction-driven {animatable} 3D human texture editing. Existing text-based 3D editing methods often directly apply Score Distillation Sampling (SDS). SDS, designed for generation tasks, cannot account for the defining requirement of editing -- maintaining consistency with the source avatar.
This work shows that naively using SDS harms editing, as it may destroy consistency.
We propose a modified SDS for Editing (SDS-E) that selectively incorporates subterms of SDS across diffusion timesteps. We further enhance SDS-E with spatial smoothness regularization and gradient-based viewpoint sampling for edits with sharp and high-fidelity detailing. Incorporating SDS-E into a 3D human texture editing framework allows us to outperform existing 3D editing methods.  Our avatars faithfully reflect the textual edits while remaining consistent with the original avatars. Project page: \url{https://jyzhu.top/instruct-humans/}.%
\end{abstract}


\section{Introduction}
\label{sec:intro}

\IEEEPARstart{W}{ith} the recent development of vision-language~\cite{radford2021clip,tmm24diffusion,tmm24MMGInpainting,sheng23exploring,tmm23Multimodal}, %
natural
language has emerged to become a control signal for generating and editing %
human avatars~\cite{liao2024tada,hong2022avatarclip,instructnerf2023,avatarstudio23,tmm25Text2Avatar}.  \rv{This work presents a novel method for text-guided \emph{editing of animatable} 3D human avatars}. Animatable avatars offer control over the 3D human pose, {though this adds challenges in aligning texture edits with an animation or pose model.}  Previous works %
are largely either not animatable~\cite{instructnerf2023,avatarstudio23} or not editable~\cite{liao2024tada,hong2022avatarclip,huang2023humannorm}. \rv{A recent work, TEDRA~\cite{sunagad2025tedra}, studies text-based editing of dynamic avatars via subject-specific retraining, whereas we aim to edit generic animatable avatars while preserving identity consistency.}

\begin{figure*}[t]
    \centering
    \includegraphics[width=\linewidth]{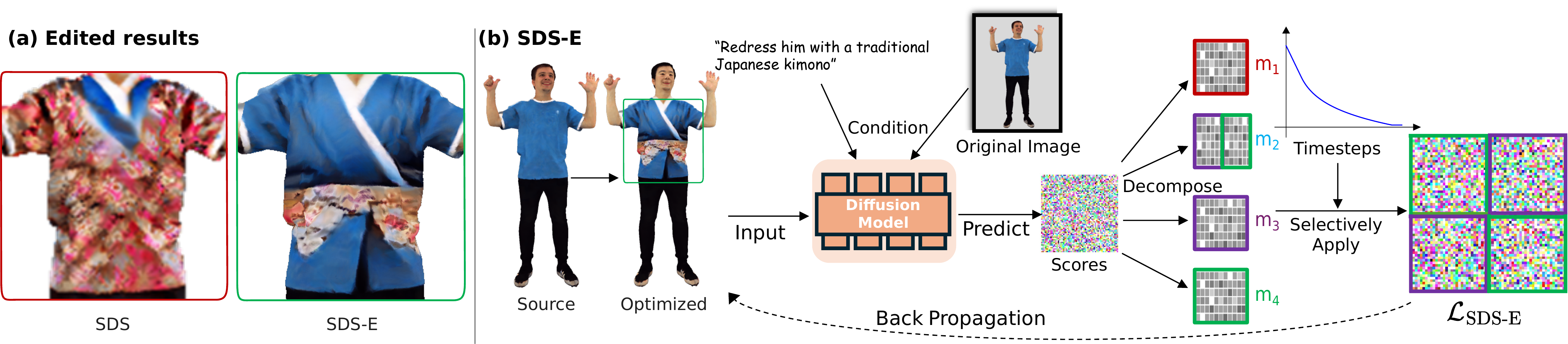}
    \caption{%
    \small \textbf{(a)} Edited results of SDS~\cite{poole_dreamfusion_2022} vs. SDS-E. SDS results in a blurred avatar, with clothing deviating from the original features.
    \textbf{(b)} SDS-E edits a human avatar by querying the diffusion model conditioned on a text instruction and the original image, decomposes predicted scores into individual terms, and selectively applies them across timesteps. By controlling these terms, SDS-E provides cleaner guidance, leading to high-quality, faithful edits while maintaining consistency in unedited regions of the avatar.
    }
    \label{fig:teaser-pipeline}
        \vspace{-.5em}
\end{figure*}

For intuitive handling of 3D avatars, 
we adapt Score Distillation Sampling (SDS)~\cite{poole_dreamfusion_2022}. SDS leverages the predicted noise of a 2D diffusion model to guide 3D model optimization. %
SDS has been effective for 3D generation~\cite{poole_dreamfusion_2022, liao2024tada, kolotouros2024dreamhuman, wang_prolificdreamer_2023, lin2023magic3d, Ruiz_2023_DreamBooth} but directly applying it to editing can lead to %
blurriness and loss of essential characteristics like facial identity or clothing details.
{Fig.}~\ref{fig:teaser-pipeline} (a) left shows an example of an avatar edited naively with SDS. It is blurry and wears a different outfit not specified in the editing text. Such drawbacks may be partially fixed by fine-tuning a personalized diffusion model~\cite{avatarstudio23} at the cost of extra compute. %

\begin{figure}[t]
    \centering
    \includegraphics[width=\linewidth]{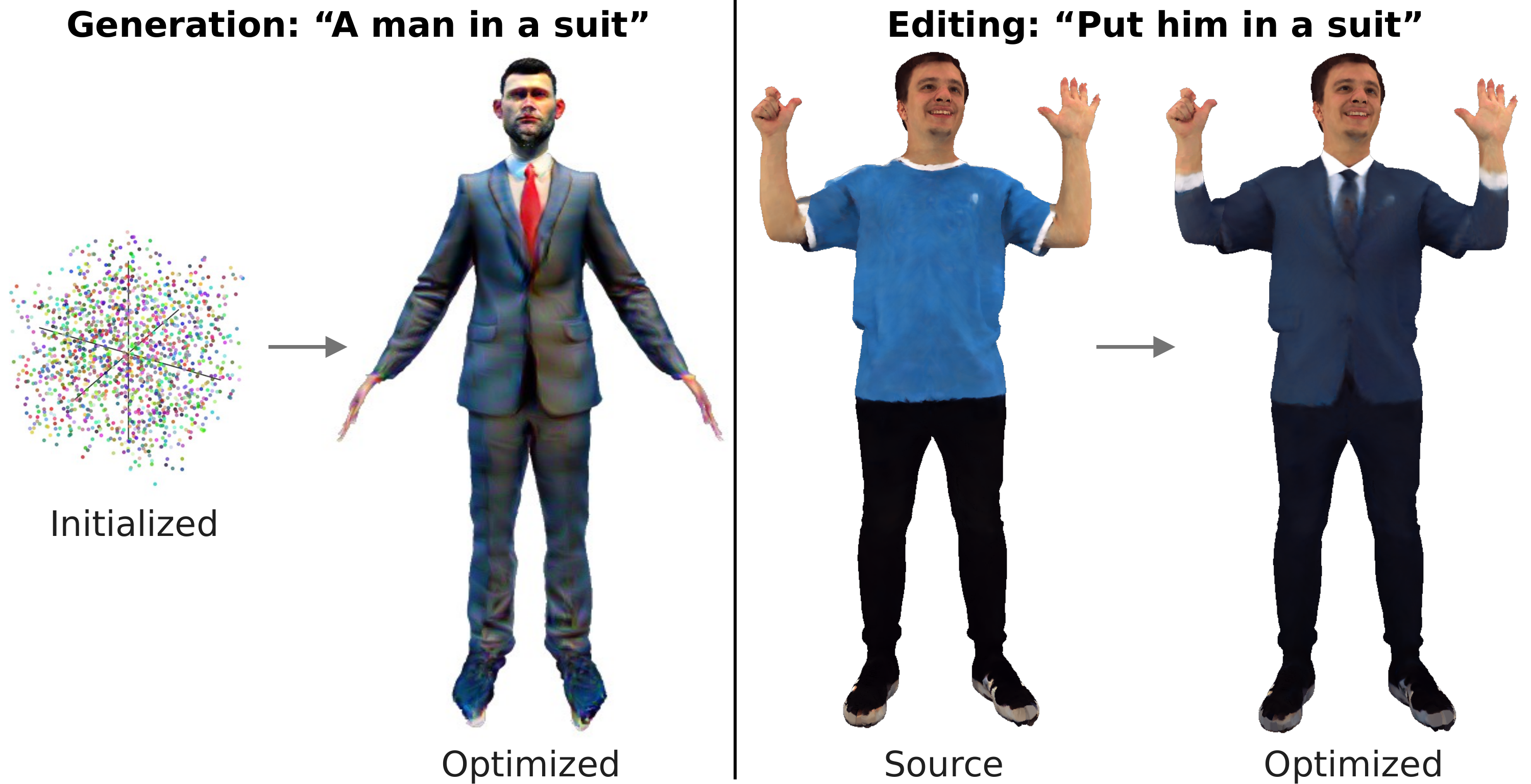}
    \caption{\small 3D avatar generation (TADA~\cite{liao2024tada}) vs. editing (Ours). 
    }
    \label{fig:teaser}
    \vspace{-1.5em}
\end{figure}

We believe the cause of the poor-quality edits is the SDS guidance signal. SDS was originally designed for generation, where the 3D model is randomly initialized.
Editing tasks, on the other hand, {begin with an existing source avatar}{ (see Fig.~\ref{fig:teaser}).}
It is necessary to preserve certain aspects of the source - %
in our case, the 3D geometry and any unaffected {facial or clothing textures} %
not specified by the edit.
This dichotomy of ``preservation'' versus ``change'' presents an inherent conflict with the guidance direction.  %

To further investigate, we break down SDS into individual terms.
Previous works~\cite{yu_csd_2023,tang2023stable} analogously showed how SDS terms affect mode-seeking and variance-reduction in the denoising process for generation. Such findings are not applicable for editing since they only consider a single condition - the generating text.
The editing scenario has two conditions - input avatar and editing text.  We also consider aspects unique to editing - specifically, how to preserve unedited features.

Our decomposition reveals that some terms are critical for structure formation in the early stages of denoising.  While meaningful for generation, they are counterproductive to editing as they cause shifts away from the original structures.  Similarly, other terms are beneficial only at later optimization stages.  If all the terms are applied naively at all stages {of denoising}, which is the case in standard SDS, the terms will conflict and have counter-productive effects that lead to poor-quality edits. Our findings motivate us to design a customized SDS for editing (SDS-E).  SDS-E distills guidance specifically for %
3D editing. %
It introduces a temporal staging that selectively applies the SDS terms, %
allowing control over the terms' impact on the editing guidance.

{To incorporate SDS-E for 3D human editing, we integrate it with a hybrid human representation~\cite{ho2023custom} and form our \textit{InstructHumans} framework. The hybrid representation uses local texture and geometry latent codes fixed to the human mesh vertices.  Such a separation allows for localized texture edits while preserving the \textit{animation} capability of edited avatars. For editing guidance, we require a 2D diffusion model conditioned on both the source image and text instruction. %
We adopt InstructPix2Pix~\cite{brooks_instructpix2pix_2023}; the only diffusion model currently supporting dual conditioning, widely used in text-based editing~\cite{instructnerf2023,Chen2023GaussianEditor,chen2024shap,Li_2024_zone}. Note, however, that SDS-E is general and can extend to other diffusion models that support dual conditioning as they emerge.}

We also investigate the spatial distribution of the distilled guidance and make two innovations that improve the efficiency and quality of the edits. First, we propose a gradient-aware view sampling strategy to allocate camera viewpoints based on the need for guidance dynamically. This strategy directs the editing focus toward desired regions and speeds up the overall editing convergence.  Secondly, we propose a smoothness regularizer to improve spatial coherence and mitigate spotting and other artifacts in the resulting textures.  

To summarize our contributions, we perform (1) an in-depth analysis of SDS for 3D editing and reveal the changing roles of the different SDS terms in the denoising process.  Based on our analysis, we introduce (2) SDS for Editing with %
selective temporal staging of the SDS terms to distill effective editing guidance.  We further introduce (3) a gradient-aware camera %
sampling that improves editing efficiency and {specificity} and (4) a smoothness regularizer that enhances the texture quality. %
Our resulting framework %
is efficient and flexible, yielding high-fidelity and faithful edits while maintaining consistency with the original avatar. %

\section{Related Works}

\noindent\textbf{Text-guided 3D Editing.}  Traditional 3D editing typically require explicit visual guidance, such as 3D cages~\cite{jambon2023nerfshop} and masks~\cite{sun2022ide}.
Recent works~\cite{brooks_instructpix2pix_2023, rombach2021highresolution, saharia2022photorealistic} try to edit 3D objects via text guidance. One line of works adopts CLIP-based similarity to guide the 3D editing~\cite{aneja2023clipface, hong2022avatarclip, jain2022dreamfield}.
The outputs are, however, unrealistic and often require additional fine-tuning with \eg GANs. Another line of work uses SDS~\cite{poole_dreamfusion_2022} to distill information from 2D diffusion models. 
Using the predicted noise to guide 3D model updates is %
practical and efficient. %
As such, SDS has been adopted by many recent works for both text-to-3D generation~\cite{li_sweetdreamer_2023, tang_dreamgaussian_2023, yi_progressive_2023, liao2024tada} and 3D editing~\cite{instructnerf2023, Hertz_2023_dds,kamata_i323_2023, yu_painthuman_2023,zhang2023teca,avatarstudio23}, applied to both scenes and avatars.
\rv{Classifier-free guidance~\cite{ho2022classifierfree} is typically adopted, and SDS~\cite{poole_dreamfusion_2022} then distills the resulting conditional and unconditional signal pairs into 3D updates. IP2P~\cite{brooks_instructpix2pix_2023} provides dual conditioning on the source image and text prompt; integrating such dual-conditioned editors within SDS for 3D optimization has seen limited exploration.}

\noindent\textbf{Improving SDS.}  %
\cite{huang_dreamtime_2023} proposes non-increasing timestep sampling; %
\cite{yu_csd_2023} proposes using only classifier guidance. \cite{tang2023stable} and \cite{katzir2024nfsd} take similar approaches as us and decompose SDS to improve stability. %
The findings of~\cite{yu_csd_2023,tang2023stable,katzir2024nfsd} improve SDS for generation, but are not applicable for editing.   %
A key limitation is that it does account for %
the preservation of features of the source avatar during optimization. 
\cite{Hertz_2023_dds} tackles this difference in 2D image editing by additionally estimating the score of the original image-text pair.
{Other works~\cite{kamata_i323_2023, yu_painthuman_2023, zhang2023teca, avatarstudio23} directly apply SDS to 3D editing in its original form, without %
modifying the individual terms, %
which leads to suboptimal results.}

\noindent\textbf{3D Human Editing.}
\rv{Methods that \emph{generate} controllable 3D humans~\cite{hong2022avatarclip, cao2023dreamavatar, kolotouros2024dreamhuman, liao2024tada} primarily optimize a generative objective and are not designed to edit an existing personal avatar. Some works, like TADA~\cite{liao2024tada} and HumanNorm~\cite{huang2023humannorm}, can appear ``edit-like'' by prompt changes, but they remain fundamentally generative, and depend on text-encoder familiarity with subjects, rather than editing a given personal avatar. \emph{Editing}, in contrast, starts from an existing human and seeks text-aligned changes while preserving identity and structure across pose and animation. Several 3D editing approaches apply SDS in its original form~\cite{kamata_i323_2023, yu_painthuman_2023, zhang2023teca, avatarstudio23}, which may be limiting for fidelity and animation consistency.
In terms of scope, many works target specific regions such as the head~\cite{aneja2023clipface, FaceCLIPNeRF23, tmm25ICE} or upper body~\cite{zhang2023teca}. %
By representation, implicit NeRF-based editors, such as Instruct-NeRF2NeRF (IN2N)~\cite{instructnerf2023} and NeRF-Art~\cite{wang2023nerfart} offer multi-view coherence but afford less direct control over topology or large deformations, while explicit mesh approaches~\cite{Michel_2022_text2mesh,kim2023chupa} provide surface-level edits under fixed topology.
Hybrid representations seek to balance these trade-offs.  EditableHumans~\cite{ho2023custom} leverages a parametric human mesh model %
and merges it %
with the adaptability and editing versatility of NeRFs. Building on this hybrid model, our work introduces text-driven editing for animatable humans, offering intuitive control and flexible representations.  
A concurrent work, TEDRA~\cite{sunagad2025tedra}, also examines text-based editing of animatable avatars; our scope differs by focusing on generic animatable avatars without per-subject model personalization.}

\section{Score Distillation Sampling for Editing} %

\subsection{Preliminaries}\label{sec:naive_sds}

\noindent\textbf{InstructPix2Pix (IP2P)}~\cite{brooks_instructpix2pix_2023} is a %
text-driven image-editing diffusion model. IP2P edits source image $I$ according to text-instructions $y$ by iteratively reducing estimated noise $\hat \epsilon_\phi$ from a noisy latent representation of the image $z=\mathcal{E}(I)$.  To that end, IP2P optimizes the following objective:
\begin{equation}
\mathcal{L}_{\text{IP2P}}(z, t, y, I) = w(t)\| \hat\epsilon_{\phi}(z_t, t, y, I) - \epsilon \|_2^2.
\end{equation}
\noindent Above, $t \in T$ denotes a uniform randomly sampled timestep, $\epsilon \sim \mathcal{N} (0, \mathbf I)$ is the ground truth Gaussian noise and $w(t)$ is a weighting function depending on $t$.  $z_t$ is the noisy latent at timestep $t$  %
generated through an iterative forward diffusion process: $z_t = \sqrt{\alpha_t}z + \sqrt{1 - \alpha_t}\epsilon$, where the coefficient $\alpha_t$ represents a predefined noise schedule.

\noindent\textbf{Classifier-free Guidance (CFG)}~\cite{ho2022classifierfree} adjusts the diffusion model's adherence to specified conditions through hyperparameter tuning.
For a %
model like IP2P with conditions $I$ and $y$, CFG is expressed as a conditional probability based on %
$I$ and $y$, with hyperparameters $\omega_t$ and $\omega_I$, respectively:
\begin{equation}
\label{eq_text&img_cond_cfg}
\begin{split}
    \hat{\boldsymbol{\epsilon}}_\phi^{CFG}(z_t, t, y, I)  = 
    \, &\hat{\mathbf{\boldsymbol{\epsilon}}_\phi}(z_t, t, \emptyset, \emptyset)  \\
    + \omega_I\cdot (&\hat{\boldsymbol{\epsilon}}_\phi(z_t, t, \emptyset, I) - \hat{\mathbf{\boldsymbol{\epsilon}}_\phi}(z_t, t, \emptyset, \emptyset)) \\
    + \omega_t\cdot (&\hat{\boldsymbol{\epsilon}}_\phi(z_t, t, y, I) - \hat{\mathbf{\boldsymbol{\epsilon}}_\phi}(z_t, t, \emptyset, I)).
\end{split}
\end{equation}
\noindent\textbf{Score Distillation Sampling (SDS)}~\cite{poole_dreamfusion_2022} leverages pre-trained 2D image diffusion models to facilitate 3D generation. %
{By applying the denoising process of IP2P %
to a rendered image from a 3D model, SDS can be used to distill editing guidance from the diffusion model into a 3D model.}
Specifically, SDS assumes that the diffusion model's noise $\boldsymbol\epsilon$ correlates with the score function (the gradient of the log-density) of the perturbed data distribution~\cite{ho2020denoising}:
\begin{equation}\label{eq:p_phi} \!\! \hat{\boldsymbol\epsilon}_\phi = -\sigma_t \nabla_{z_t}\log p(z_t;t,y,I), \;\; \text{where } \sigma_t=\sqrt{1\!-\!\alpha_t}.
\end{equation}
\noindent This assumption means SDS directs updates towards the data distribution $p(z_t)$'s high-density regions.
Applied to a 3D model parameterized by $\Theta$, the gradient is given as:
\begin{equation}\label{eq:van_sds}
    \nabla_\Theta \mathcal{L}_{SDS}(\phi, z)=[w(t)(\hat{\boldsymbol{\epsilon}}_\phi^{CFG}(z_t; t,y,I) - \boldsymbol{\epsilon}) \frac{\partial z_t}{\partial \Theta}],
\end{equation}
where $\hat{\boldsymbol{\epsilon}}_\phi^{CFG}(z_t; t,y,I)$ is the IP2P model's noise estimation guided by CFG.

\noindent\rv{\textbf{Decomposition of SDS (Generation Setting).}
For single-condition (text-only) diffusion models used for 3D generation, SSD~\cite{tang2023stable} reorganizes the SDS guidance into two subterms:
\begin{equation}
\label{eq:ssd_single}
\begin{split}
    \hat{\boldsymbol{\epsilon}}_\phi^{CFG}(\boldsymbol{x}_t; t, y) - \boldsymbol{\epsilon}
    &= \omega\cdot\underbrace{\bigl(\hat{\boldsymbol{\epsilon}}_\phi(z_t, t, y) - \hat{\boldsymbol{\epsilon}}_\phi(z_t, t, \emptyset)\bigr)}_{\text{mode-disengaging}} \\
    &\quad+ \underbrace{\hat{\boldsymbol{\epsilon}}_\phi(z_t, t, y) - \boldsymbol{\epsilon}}_{\text{mode-seeking}}.
\end{split}
\end{equation}
From the score view (Eq.~\ref{eq:p_phi} with a single condition), the mode-disengaging term compares $\nabla\log p(z_t;t,y)$ against $\nabla\log p(z_t;t)$ and, at small $t$, maximizes $\frac{p(z;y)}{p(z)}$. This tends to decrease $p(z)$ and leads to saturation. SSD therefore recommends omitting the corresponding contribution at small timesteps; the mode-seeking term, when used alone with uniform $t$, can trap optimization in intermediate modes, contributing to over-smoothing. We recall these behaviors here as background; our analysis below adapts the decomposition and implications to the dual-conditional (text and image) editing setting.}

\noindent{\textbf{3D Generation vs. Editing.}}\label{sec:gen_vs_edit}
Advancements in extending 2D diffusion models to 3D using techniques like SDS~\cite{poole_dreamfusion_2022} have led to two main tasks: \textit{3D generation}~\cite{poole_dreamfusion_2022,tang2023stable,katzir2024nfsd,yu_painthuman_2023,hong2022avatarclip,wang_prolificdreamer_2023} and \textit{3D editing}~\cite{instructnerf2023,Hertz_2023_dds}.

In \textit{3D generation}, the goal is to synthesize a 3D representation $\Theta_{\text{tgt}}$ from an initial random state $\Theta_0$, guided by a textual prompt $y$:
\begingroup
\begin{equation}\label{eq:generation}
    \Theta_{\text{tgt}} = \operatorname*{arg\,min}_{\Theta} \ \mathcal{L}_{\text{gen}}(\Theta, y), \, \text{starting from } \Theta = \Theta_0,
\end{equation}
\endgroup
where $\mathcal{L}_{\text{gen}}(\Theta, y)$ measures how well $\Theta$ aligns with $y$, such as using the SDS loss.

In contrast, \textit{3D editing} transforms an existing 3D representation $\Theta_{\text{ori}}$ into $\Theta_{\text{tgt}}$, based on an editing instruction $y$, while preserving original content. It utilizes images rendered from $\Theta_{\text{ori}}$, denoted as $I = \mathcal{R}(\Theta_{\text{ori}})$, where $\mathcal{R}$ is the rendering function: %
\begingroup
\begin{equation}\label{eq:editing}
\!\!\!\!    \Theta_{\text{tgt}} = \operatorname*{arg\,min}_{\Theta} \ \mathcal{L}_{\text{edit}}(\Theta, \mathcal{R}(\Theta_{\text{ori}}), y), \,\text{starting from } \Theta = \Theta_{\text{ori}},\!\!\!\!
\end{equation}
\endgroup
where $\mathcal{L}_{\text{edit}}(\Theta, \mathcal{R}(\Theta_{\text{ori}}), y)$ measures alignment with $y$ while retaining features of $\Theta_{\text{ori}}$.
Computing $\mathcal{L}_{\text{edit}}$ requires a diffusion model conditioned on both image and text; we adopt IP2P~\cite{brooks_instructpix2pix_2023}, {which supports dual conditioning and is widely used~\cite{instructnerf2023,Chen2023GaussianEditor,chen2024shap,Li_2024_zone}}. %
Yet, our analysis and method are general and can be extended to other diffusion models that support dual-conditioning if available.

Our work focuses on 3D editing. Despite sharing common aspects like being text-based and SDS decomposition, 3D generation works~\cite{poole_dreamfusion_2022,tang2023stable,katzir2024nfsd,yu_painthuman_2023,hong2022avatarclip,wang_prolificdreamer_2023} are not directly comparable due to the differences in the objectives outlined in Eq.~\ref{eq:generation} vs. Eq.~\ref{eq:editing}.  {It is also worth noting that some works like TADA}~\cite{liao2024tada} and HumanNorm~\cite{huang2023humannorm} extend their method to an edit-like setting.  However, they are fundamentally generative, as they follow the same objective as Eq.~\ref{eq:generation}.  %
They `edit' by altering keywords in the generation prompts (\eg changing ``Messi in a suit'' to ``Messi in a jacket.'') This strategy relies on the text encoder's familiarity with specific subjects (in this case Messi), %
and does not extend to arbitrary individuals like our work. %

\subsection{Score Distillation Sampling for Editing}\label{sec:analysis}
\noindent\textbf{Timestep Sampling.} Standard SDS samples timesteps $t$ uniformly at random, but prior analysis shows that large timesteps are crucial for forming coarse features, while middle\footnote{Timestep sizing is relative. To facilitate our discussion, we separate ``small'', ``middle'' and ``large'' timesteps, though our ``small'' and ``middle'' correspond to the ``small'' timesteps of~\cite{tang2023stable}, as they only make a distinction between ``small'' and ``large''.} and small timesteps are geared towards detailing~\cite{choi2022perception}.
In the context of 3D editing, a source 3D representation exists. Large timesteps serve little value and even risk disrupting the original structure, so we opt to fully remove large timesteps from the sampling.  

Previously,~\cite{huang_dreamtime_2023} proposed a non-increasing timestep sampling strategy which they showed to be more informative for updating 3D neural fields. %
The sampling strategy enforces a monotonically decreasing envelope function to ensure that sampled timesteps are non-increasing.  We observe that using this sampling strategy for our 3D human editing is more effective, as the successively smaller timesteps facilitate the escape of intermediate modes and promote convergence towards the optimal edited mode.

\noindent\textbf{Decomposition of Dual-Conditioned SDS.}\label{sec:decomp_sds}
Our analysis begins with decomposing SDS for a dual-conditional diffusion model. This process aims to distinguish the editing directions influenced by the conditions and those influenced by the baseline (unconditioned) noise model. Adapting Eq.~\ref{eq:ssd_single} to two conditions $(y,I)$ by substituting Eq.~\ref{eq_text&img_cond_cfg} into Eq.~\ref{eq:van_sds} yields:
\begin{equation}
\scriptsize
\label{eq:sds_decompose_1}
\begin{split}
  \!\!  &\hat{\boldsymbol{\epsilon}}_\phi^{CFG}(z_t; t, y, I) - \boldsymbol{\epsilon} 
    = (\omega_I\!-\! 1)\cdot\underbrace{\bigl(\hat{\boldsymbol{\epsilon}}_\phi(z_t, t, \emptyset, I) -  \hat{\boldsymbol{\epsilon}}_\phi(z_t, t, \emptyset, \emptyset)\bigr)}_{m_1} \\
  \!\! &+  \underbrace{\omega_t\cdot \bigl(\hat{\boldsymbol{\epsilon}}_\phi(z_t, t, y, I) - \hat{\boldsymbol{\epsilon}}_\phi(z_t, t, \emptyset, I)\bigr) +  \hat{\boldsymbol{\epsilon}}_\phi(z_t, t, \emptyset, I)
   - \boldsymbol{\epsilon}}_{m_2}.\!\!
\end{split}
\end{equation}
\noindent The first part, $m_1$,
weighted by $\omega_I - 1$, is a  \textit{baseline-shift term}.  $m_1$ quantifies the %
divergence induced by the image condition $I$, since it measures the shift from a baseline (unconditioned) noise model to a conditionally influenced model.  Note this term measures shift from $I$ only, and does not account for the text instruction.  The second part, $m_2$, is a \textit{condition-integration term}, as it integrates the condition of the text instruction $y$ and helps align the generated output with both conditions of $I$ and $y$. %

Since $m_2$ involves both conditions, it can be further re-arranged into a form analogous to Eq.~\ref{eq:sds_decompose_1}: 
\begin{equation}\small
\label{eq:sds_decompose_2}
\begin{split}
    m_2
   = (\omega_t - 1)\cdot &\underbrace{(\hat{\boldsymbol{\epsilon}}_\phi(z_t, t, y, I) - \hat{\boldsymbol{\epsilon}}_\phi(z_t, t, \emptyset, I))}_{m_3}  \\
   &+\underbrace{\hat{\boldsymbol{\epsilon}}_\phi(z_t, t, y, I)
   - \boldsymbol{\epsilon}}_{m_4}.
\end{split}
\end{equation}\vspace{-.2em}
\noindent The term 
$m_3$, weighted by $\omega_t - 1$, is a \textit{condition-divergence term} that measures the adjustment needed when shifting from a base image condition to integrate the text condition $y$.  Meanwhile, $m_4$ is the \textit{full-condition term}, as it captures the model's output with full consideration of the conditions.

\noindent\textbf{Analysis of the Baseline-Shift Term $m_1$.}
Replacing the unconditional reference in the SSD~\cite{tang2023stable} argument with the image-conditioned reference yields:
\begin{equation}
\begin{split}
m_1 &= \hat{\boldsymbol{\epsilon}}_\phi(z_t, t, \emptyset, I) - \hat{\boldsymbol{\epsilon}}_\phi(z_t, t, \emptyset, \emptyset) \\
    &= -\sigma_t(\nabla_{z_t}\log p_\phi(z_t;t,I) - \nabla_{z_t}\log p_\phi(z_t;t)). 
\end{split}
\end{equation}
The term $m_1$ causes shifts away from natural image distributions at small (and middle) timesteps. Specifically, when $t\rightarrow 0$, the distributions $p_\phi(z_t;t,I) \rightarrow p_\phi(z;t,I)$ and $p_\phi(z_t;t)\rightarrow p_\phi(z;t)$ lead to the maximization of the following term:
\begin{equation}
     \!\!\frac{p_\phi(z_t;t,I)}{p_\phi(z_t;t)} 
     \rightarrow \frac{p_\phi(z;t,I)}{p_\phi(z;t)} = \frac{p_\phi(z;I)}{p_\phi(z)}.\!\!
\end{equation}
Therefore, as in SSD's single-condition case, we omit $m_1$ for small (and middle) timesteps.

\noindent\textbf{Analysis of the Condition-Divergence Term $m_3$.}
Similar to $m_1$, $m_3$ is characterized by two directional influences: one toward the two conditional mode $p_\phi(z_t;t,y,I)$ and one away from the image conditional mode $p_\phi(z_t;t,I)$.  Again, in the limit $t\!\rightarrow\! 0$, the condition-divergence term maximizes:
\begin{equation}
\label{eq:lim_m3}
     \frac{p_\phi(z_t;t,y,I)}{p_\phi(z_t;t,I)} 
     \rightarrow \frac{p_\phi(z;t,y,I)}{p_\phi(z;t,I)} = \frac{p_\phi(z;y,I)}{p_\phi(z;I)}. 
\end{equation}
Analogous to $m_1$, we assume that a mode of $p(z;y,I)$ should also be a mode of $p(z;I)$.  Yet Eq.~\ref{eq:lim_m3} cannot effectively drive the editing process towards a maximum of $p_\phi(z;y,I)$ where $p_\phi(z;I)$ is also high, as the latter term sits in the denominator and gets minimized during the optimization process. This discourages convergence to any significant mode of the distribution $p_\phi(z;I)$, and distances the result from the original image. 

While $m_3$ should presumably be removed at \emph{small} timesteps, empirical evidence suggests that it significantly guides the text-conditioned mode and \emph{improves} alignment with instructions. This allows for greater control over the trade-off between editing faithfulness and image fidelity—two competing factors central to editing tasks, which differ from single-objective generation. Therefore, we regard the inclusion of $m_3$ as a flexible design choice, reflecting a balance between these two aspects.
In principle, $m_3$ should also be removed for \emph{middle} timesteps; however, this would leave only the $m_4$ term for guidance, which is problematic in its own right.  %
We further elaborate in the analysis on $m_4$.

\noindent\textbf{Analysis of the Full-Condition Term $m_4$.}\label{sec:interm_trap}
The full-condition term $m_4$ can be viewed as a guide towards a two-condition mode $p_\phi(z_t;t,y,I)$.  It is augmented by a factor $-\boldsymbol{\epsilon}$ that counterbalances the variance introduced by the noise without altering the targeted mode:
\begin{equation}
\begin{split}
m_4 & = \hat{\boldsymbol{\epsilon}}_\phi(z_t, t, y, I) - \boldsymbol{\epsilon} \\
    & = -\sigma_t\nabla_{z_t}\log p_\phi(z_t;t,y,I) - \boldsymbol{\epsilon}. 
\end{split}
\end{equation}
Applying $m_4$ alone may trap the model in intermediate modes.  %
In particular, for large or middle timesteps, {denoising is incomplete}, %
{so the peak of a joint probabilistic density with multiple modes is likely higher than that of any individual desired mode~\cite{tang2023stable}.} This issue diminishes in smaller timesteps, when the probabilistic density of the desired mode becomes higher, dominating the update direction.  Yet in a uniformly random timestep sampling strategy, as in standard SDS, %
revisiting large or middle timesteps allows this issue to persist and disrupts convergence to any desired mode. This is the root cause for over-smoothing by SDS~\cite{tang2023stable,poole_dreamfusion_2022}.
\begin{table}[t!]
    \caption{\small Impact of SDS terms at different timesteps. The shading indicates utility; red and green denote harmful and helpful respectively, while yellow denotes mixed effects. %
    }
    \label{tab:sdse_components}
    \centering
    \vspace{-0.5em}
    \scriptsize
    \begin{tabular}{|c|ccc|}
    \hline
Timestep Sizing & \multicolumn{1}{c|}{$m_1$}                                                & \multicolumn{1}{c|}{$m_3$}                                       & $m_4$                                     \\ \hline
Large ($>800$)         & \multicolumn{3}{c|}{\cellcolor[HTML]{FFCCC9}Counterproductive}                                                                                                                           \\ \hline
Middle (150-800)        & \multicolumn{1}{c|}{\cellcolor[HTML]{FFCCC9}}                             & \multicolumn{1}{c|}{\cellcolor[HTML]{FFFC9E}Counterbalance}                & \cellcolor[HTML]{FFFC9E}Intermediate trap \\ \cline{1-1} \cline{3-4} 
Small ($<150$)         & \multicolumn{1}{c|}{\multirow{-2}{*}{\cellcolor[HTML]{FFCCC9}Saturation}} & \multicolumn{1}{c|}{\cellcolor[HTML]{FFFC9E}Distant from image} & \cellcolor[HTML]{9AFF99}Two condition     \\ \hline
    \end{tabular}
    \vspace{-1em}
\end{table}

As such, we can either %
remove $m_4$ at middle timesteps and use only $m_3$, or combine $m_3$ and $m_4$ (\ie keep the full $m_2$ term).  Empirically, the latter is better. Using $m_3$ alone shifts the output too far towards the text-conditioned mode and is problematic to optimize in its own right, as we analyzed previously.  Using the two together allows $m_4$ to facilitate a balance of the text and image conditions while allowing $m_3$ to provide a counterbalance for breaking free of intermediate modes. Combining $m_3$ and $m_4$ with non-increasing timestep sampling~\cite{huang_dreamtime_2023} produces the best results.

\noindent\textbf{SDS-E: Score Distillation Sampling for Editing.} \label{sec:sdse}
Our analysis of the three SDS terms based on timestep size is summarized in  Tab.~\ref{tab:sdse_components}.  Based on these findings, we present a customized SDS for editing (SDS-E), where we selectively apply the terms at distinct timestep sizes.   %

For each sampled timestep $t$, SDS-E is defined as:
\begin{equation}
\begin{split}
    \mathcal{L}_\text{SDS-E}= \omega_t\cdot ( &\hat{\boldsymbol{\epsilon}}_\phi(\boldsymbol{x}_t, t, y, I) - \hat{\boldsymbol{\epsilon}}_\phi(\boldsymbol{x}_t, t, \emptyset, I)) \\
    + & \hat{\boldsymbol{\epsilon}}_\phi(\boldsymbol{x}_t, t, \emptyset, I)
   - \boldsymbol{\epsilon}.
\end{split}
\end{equation}
We also consider an alternative where the condition-divergence term $m_3$ is excluded at small timesteps:
\begin{equation}
\!\!\!    \mathcal{L}'_\text{SDS-E}=
    \begin{cases}
   \mathcal{L}_\text{SDS-E}
   & \text{if } t > M \\
        \hat{\boldsymbol{\epsilon}}_\phi(\boldsymbol{x}_t, t, y, I) - \boldsymbol{\epsilon} & \text{if } t \leq M,
    \end{cases}
\end{equation}
where $M$ is the threshold between small and middle timesteps. We empirically set $M$ to 150 and limit middle timesteps to a maximum of $800$ to exclude larger timesteps. %

\section{InstructHumans Editing Pipeline}

\begin{figure}[t!]
    \centering
    \includegraphics[width=\linewidth]{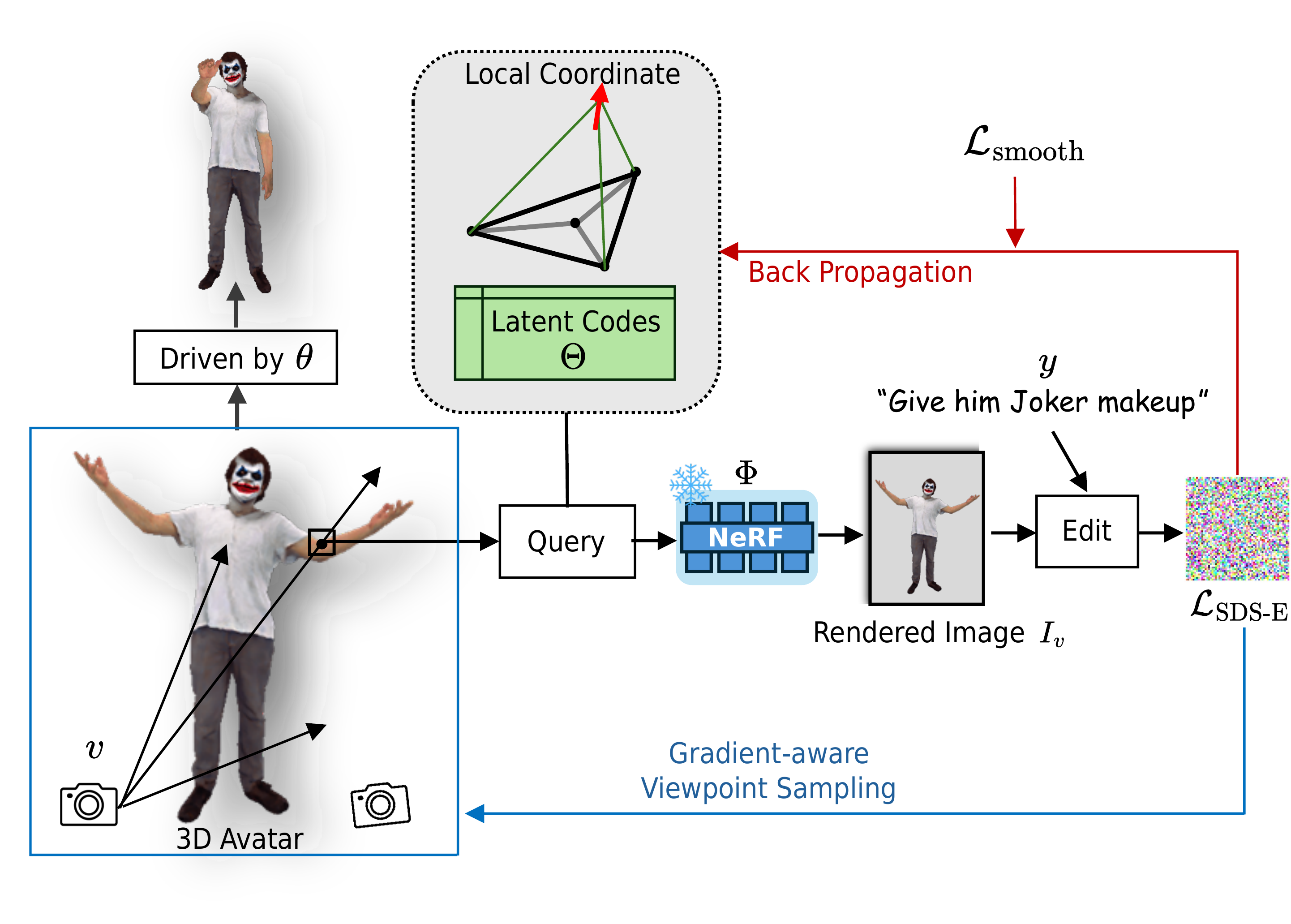}

    \vspace{-2.5em}
    \caption{\small \textbf{Instruction-driven 3D human editing pipeline.} Our pipeline optimizes a specific human subject's texture based on textual instructions. Images rendered through a conditional NeRF are edited by IP2P, with SDS-E used to distill the editing gradients and update the texture latent codes. %
    The editing is enhanced by gradient-aware viewpoint sampling and a smoothness regularizer. The edited avatar is easily drivable by altering pose parameters.}%
    \label{fig:method}
    \vspace{-0.5cm}
\end{figure}

\noindent\textbf{Hybrid 3D Human Representation.}
We adopt the hybrid 3D human representation proposed by EditableHumans~\cite{ho2023custom}. It associates an explicit %
3D human mesh model, SMPL-X~\cite{SMPL-X-2019}, with an implicit NeRF.  Each mesh vertex from SMPL-X is linked with local geometry and texture latent codes. For a specific 3D avatar,
it stores trainable latent codes $\Theta$, obtained by barycentric interpolation of three local features accessed via vertex indices. 
EditableHumans also contains a pre-trained NeRF or implicit network, which outputs RGB color $c$ and SDF value $s$ for any queried global coordinate $x_g$. Specifically, the implicit network $\mathbf{\Phi}$ is provided with a local coordinate $x_l = \mathcal{M}(x_g)$ that is transformed from the global coordinate $x_g = (x, y, z)$ and a local normal vector $\overrightarrow{n}$. Conditioned on the latent codes $\Theta$, the implicit network provides the following:
\begin{equation}
    \mathbf{\Phi}(\Theta, x_l, \overrightarrow{n}) = \Bigl(c(x_g), s(x_g)\Bigr).
\end{equation}
The global to local coordinate transformation finds the nearest triangle on the body mesh of an input query point $x_g$ and transforms the position into local triangle coordinates $x_l=(u, v, d)$, where $d$ is the distance. $\overrightarrow{n}$ is calculated as the direction from the closest point on the mesh to the global position, providing auxiliary positional information.
This transformation ensures that the NeRF accesses only local features, prevents it from memorizing global information, and disentangles local features for further editing.

\noindent\textbf{Editing Pipeline} (see Fig.~\ref{fig:method}). %
Starting with an input human subject with pre-trained latent codes $\Theta$ at mesh vertices, we optimize the latent codes %
to modify the human texture. At each iteration, an image $I_v$ from a sampled camera view $v$ is rendered with a conditional NeRF $\mathbf{\Phi}$. Image $I_v$ is provided to IP2P for editing, conditioned on the instruction $y$ and an original image rendered from the same view. We use our proposed SDS-E to distill editing gradients from IP2P (Sec.~\ref{sec:sdse}). The gradients, together with a smoothness regularizer $\mathcal{L}_\text{smooth}$ (see Eq.~\ref{eq:smoothness}), are backpropagated for optimizing the latent codes. Gradient-aware viewpoint sampling dynamically adjusts the camera views based on the gradients (see Eq.~\ref{eq:view-count}). The edited human is easily drivable by changing the SMPL-X pose parameter $\theta$.

\noindent\textbf{Laplacian Smoothness Regularization.}
SDS-based 3D optimization often suffers from high-frequency noise due to three factors:  
(1) Randomly sampled camera views provide inconsistent supervision for overlapping 3D regions, leading to multi-view inconsistency~\cite{poole_dreamfusion_2022}.  
(2) The distilled guidance is inherently noisy due to network instability~\cite{katzir2024nfsd} or architectural limitations~\cite{pan2024pgc}.  
(3) In discrete parameterizations (\eg per-vertex latent codes~\cite{ho2023custom} or 3D Gaussians~\cite{kerbl3Dgaussians}), SDS gradients update each location independently, amplifying noise and causing texture artifacts (Fig.~\ref{fig:ablation}).  

We observe that guidance at a 3D location should be consistent across both views and adjacent neighbors. Given the explicit connectivity of the mesh, we introduce Laplacian latent smoothness to enforce spatial coherence in SDS updates. Inspired by Laplacian constraints in surface reconstruction~\cite{sorkine2004laplacian}, we define:
\begin{equation}\label{eq:smoothness}
    \mathcal{L}_\text{smooth} = \frac{1}{N}\sum_i^N \|(\overrightarrow{L} \overrightarrow{\Delta F})_i\|^2,
\end{equation}
where $N$ is number of vertices, $\overrightarrow{L}$ is the Laplacian matrix encoding connectivity between vertices, and $\overrightarrow{\Delta F}$ is the matrix of delta latent codes, with each row representing the delta vector in latent space before and after one iteration. 
This regularizer penalizes local inconsistencies while preserving global details, improving texture coherence and convergence stability by reducing high-frequent flickering.
The overall gradient is:
\begin{equation}
    \nabla_\Theta(w_1 \mathcal{L}_\text{smooth} + \mathcal{L}_\text{SDS-E}).
\end{equation}
\noindent\textbf{Gradient-Aware Viewpoint Sampling.}
Another challenge is that edits %
are distributed unevenly across body regions depending on the text instructions. %
For example, ``Put the person in a suit'' targets clothing on the entire body, while ``Give him Joker makeup'' emphasizes the facial features.  Uniform random viewpoint sampling misallocates editing effort. %
{Some generation methods~\cite{liao2024tada, kolotouros2024dreamhuman, yu_painthuman_2023,hong2022avatarclip} simply prioritize facial views, but this is not suitable for editing since not all prompts require significant modifications to the face.}
As such, we introduce the concept of editing strength,    
defined as the average gradient magnitude anchored at a region of vertices, and  %
prioritize regions according to editing strength. %
First, we split the $10,475$ mesh vertices from SMPL-X into $5$ regions based on their source: the face, the back of the head, the front body, the back of the body, and the arms. Note that this division is flexible and can be adapted for different applications.
We then conduct one editing iteration with a batch of uniformly sampled $|V|$ views and calculate the average gradient magnitude $w_r$ across the region $r$:
\begin{equation}\label{eq:view-weight}
    w_r = \frac{1}{|V|} \frac{1}{|S_r|} \sum_{v\in V} \sum_{i\in S_r} \|\nabla (i) \|,
\end{equation}
where $V$ denotes the set of sampled views, and $S_r$ represents the set of vertices within region $r$. 
Using $w_r$ as a normalization weight, we set $\mathcal{C}(r)$ as the the total number of views sampled for region $r$ as follows:
\begin{equation}\label{eq:view-count}
    \mathcal{C}(r) = \frac{w_r}{ \sum_{r\in R}w_r} |V|,
\end{equation}
to {redistributes the number of camera views per region.}
Implementing this technique allows us to cap the number of sampled views at a predefined limit, \eg $1000$, and significantly reduce the time required for rendering. It also accelerates the convergence rate, leading to a reduction in the overall number of editing iterations needed. Moreover, it improves the editing specificity on the desired regions, facilitating editing quality.

\noindent\rv{\textbf{Selective Local Editing.}\label{sec:selective_local_editing}
As an alternative to gradient-aware viewpoint sampling, we can specify exact regions for editing by leveraging the controllability of 3D human models. 
Using the same body-region partition as above, a large language model (LLM) assistant maps an instruction to one or more target regions (represented by mesh vertices). SDS-E gradients are then applied only to the latent codes of the selected regions, leaving others unchanged. This option is useful when the intent is localized (\eg accessories or facial attributes), whereas gradient-aware sampling adapts automatically when edits are distributed across multiple regions.}

\section{Experiments}

\begin{figure*}[!htbp]
    \centering
    \includegraphics[width=\linewidth]{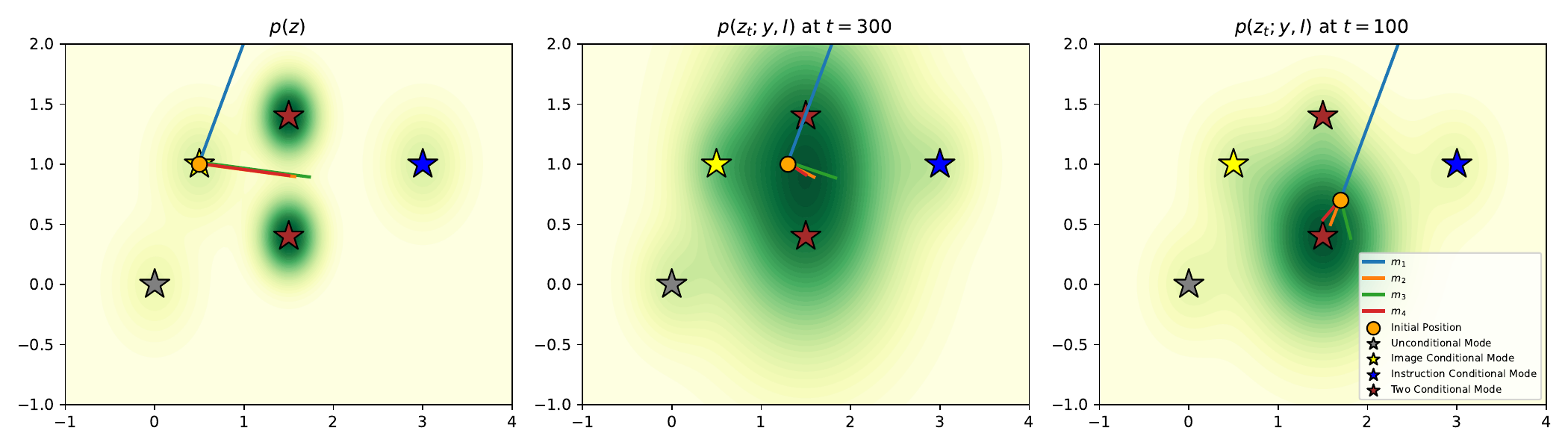}
    \caption{%
    \small \textbf{Visualizing the impact of SDS components $m_1$, $m_2$, $m_3$, and $m_4$ in a 2D toy example.} Each component, serving as an estimator, guides the optimization of $z=\theta\in \mathbb{R}^2$ within a Gaussian mixture model representing $p(z)$.  The objective is to guide $\theta$ towards the two-conditional modes (red stars). \textbf{Left}: In an early phase, $\theta$ is initiated at the image conditional mode (yellow star). The trajectory indicates that $m_1$ is counterproductive. \textbf{Center}: At middle timesteps, $m_4$ faces entrapment by an intermediate mode, while $m_3$ (and by extension, $m_2$) facilitates escape. \textbf{Right}: In small timesteps, as $\theta$ nears the target mode, $m_4$ and $m_2$ drives towards the denser region, while $m_3$ guides a deviated direction due to distancing from the image conditional mode.
    }
    \label{fig:appen-toy-example}
\end{figure*}

\subsection{Evaluating SDS Components via a Toy Example}\label{sec:appen_toy_example}

We assess the behavior of the SDS components $m_1$, $m_2$, $m_3$, and $m_4$ (Sec.~\ref{sec:decomp_sds}) using a 2D toy example. In this experiment, we optimize $z=\theta\in\mathbb{R}^2$
over a Gaussian mixture model defined as $p(z) = 0.1\,\mathcal{N}([0,0]^\top, 0.1\mathbf{I}) + 0.15\,\mathcal{N}([3,1]^\top, 0.1\mathbf{I}) + 0.15\,\mathcal{N}([0.5,1]^\top, 0.1\mathbf{I}) + 0.3\,\mathcal{N}([1.5,1.4]^\top, 0.05\mathbf{I}) + 0.3\,\mathcal{N}([1.5,0.4]^\top, 0.05\mathbf{I})$.
Here, the mode at $[0,0]^\top$ is unconditional, $[0.5,1]^\top$ is image conditional, $[3,1]^\top$ is text conditional, and $[1.5,1.4]^\top$ along with $[1.5,0.4]^\top$ denote the two-conditional modes. We simulate the 3D editing process by guiding $\theta$ toward the two-conditional modes using the estimators formulated in Eqs.~\ref{eq:sds_decompose_1} and \ref{eq:sds_decompose_2}.
Figure~\ref{fig:appen-toy-example} illustrates three learning phases that align with our analyses in Tab.~\ref{tab:sdse_components}.

\subsection{Experiment Settings}
\begin{figure*}[tbh]
    \centering
    \includegraphics[width=0.95\linewidth]{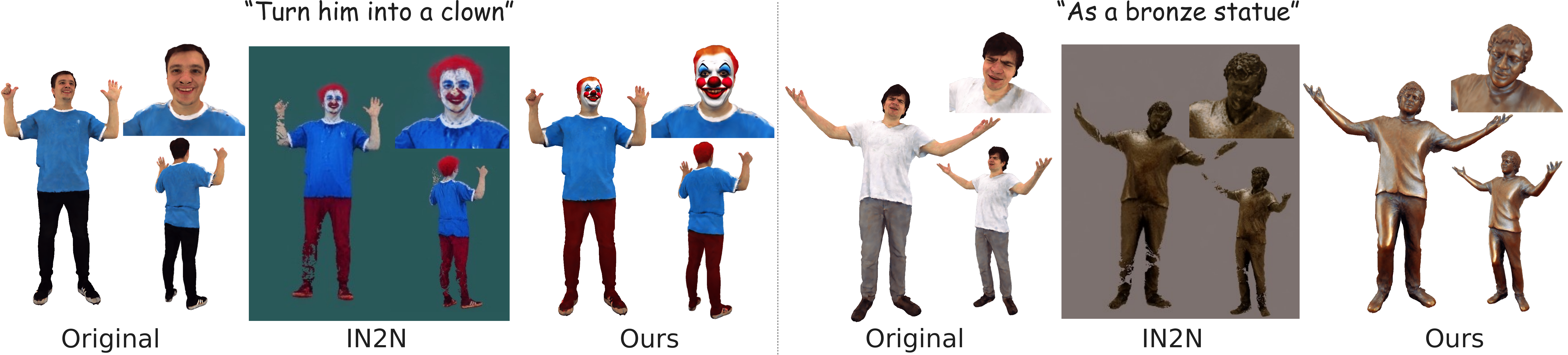}
    \caption{\small \textbf{Qualitative comparison with IN2N}.  IN2N struggles with content preservation and texture quality.}
    \label{fig:qual-comp-in2n}
\end{figure*}

\begin{figure*}[tbh]
    \centering
    \includegraphics[width=0.95\linewidth]{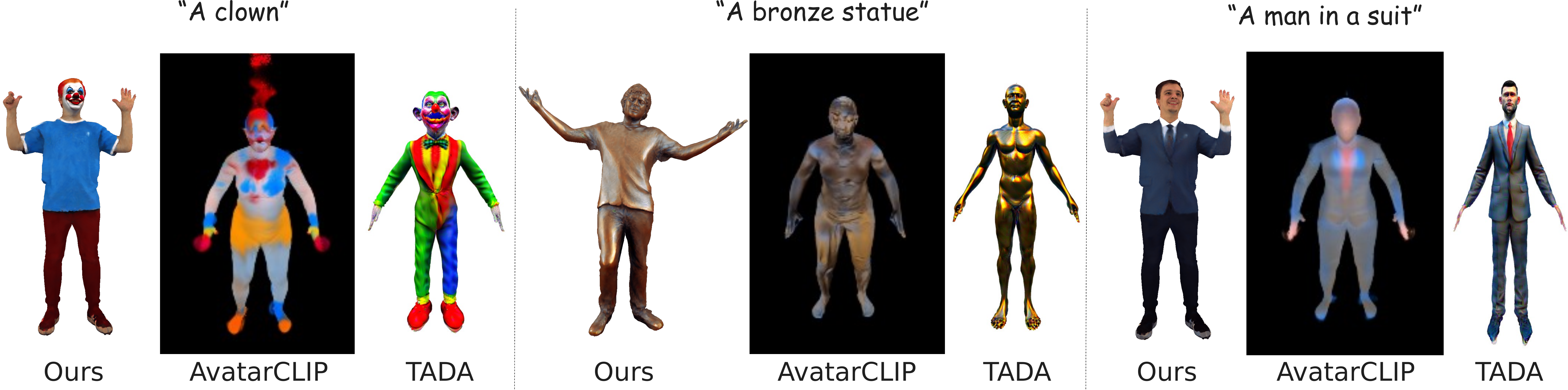}
    \caption{\small \textbf{Qualitative comparison with AvatarCLIP and TADA}. Ours achieves superior photorealistic quality.}
    \label{fig:qual-comp-avatarclip}
\end{figure*}

\begin{figure}[tbh]
    \centering
    \includegraphics[width=0.95\linewidth]{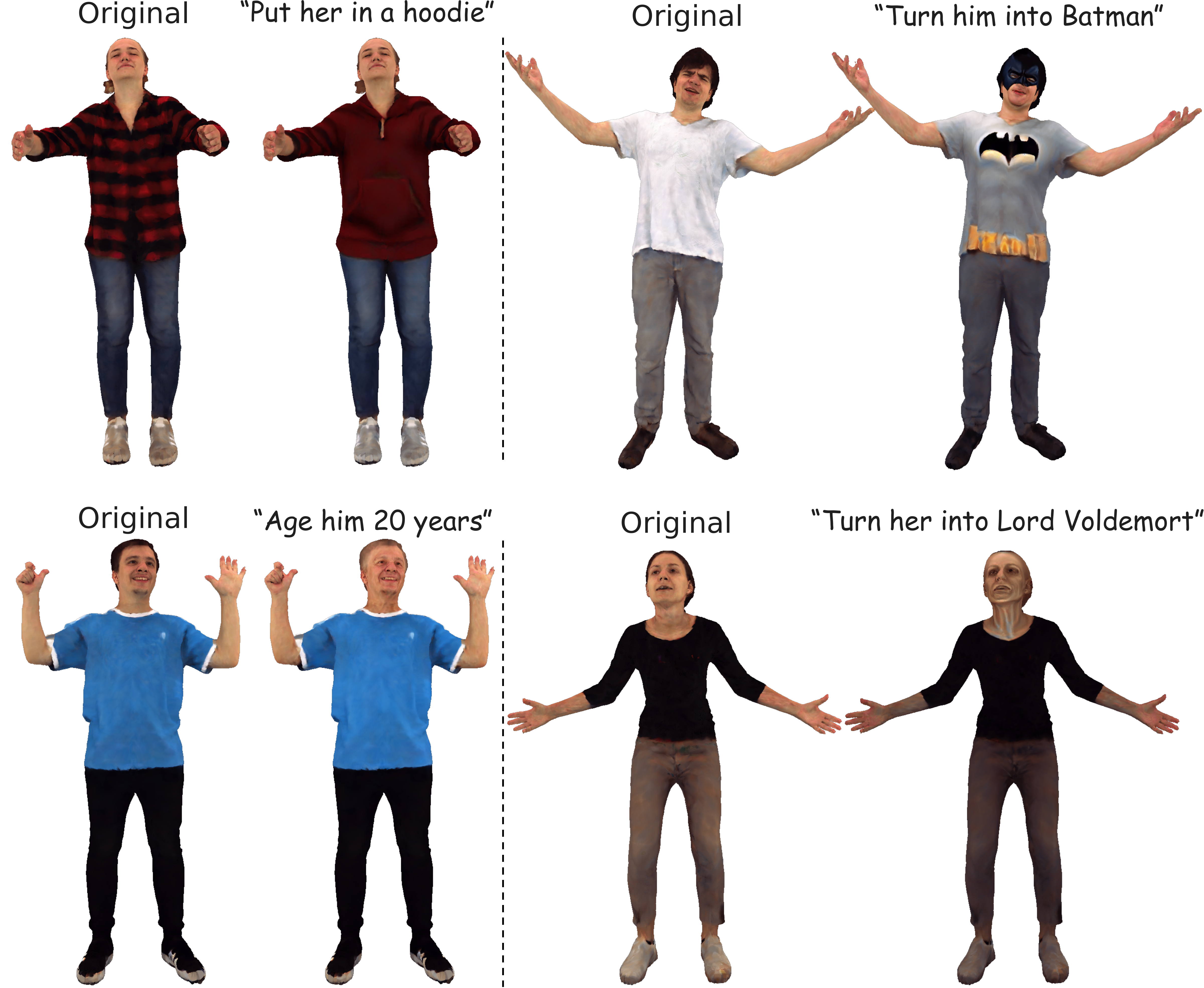}
    \caption{\small \textbf{Qualitative visualization of our results.} }
    \label{fig:qual_results}
\end{figure}

\noindent\textbf{Comparison Methods.}
Our goal is to edit animatable 3D human textures based on text instructions. Directly comparable prior work is limited;  %
we adapt related text-based methods for fair comparison, both qualitatively (Sec.~\ref{sec:qual}) and quantitatively (Sec.~\ref{sec:quan}).  

We compare with IN2N~\cite{instructnerf2023}, \rv{which edits NeRF texture/geometry but is non-animatable in its original setting; we therefore use it in static comparisons.} %
We also compare with SDS-based methods~\cite{poole_dreamfusion_2022,katzir2024nfsd,tang2023stable}. %
Their focus on 3D generation differs from our 3D editing task, precluding a direct comparison in the original framework. \rv{To isolate the editing objective while ensuring animatability, we instantiate SDS, SSD, and NFSD within EditableHumans~\cite{ho2023custom} and drive all edited avatars with identical pose/motion inputs.} Specifically, we substitute SDS-E in our pipeline with SDS, SSD, and NFSD for fair comparison.
Lastly, we compare with the avatar generation methods AvatarCLIP~\cite{hong2022avatarclip} {and TADA}~\cite{liao2024tada}; since generation and editing methods have different objectives and evaluation metrics (Sec.~\ref{sec:gen_vs_edit}), we adapt our edit prompts into generation prompts and assess the resulting avatars' quality relative to the prompts. \rv{We also visualize TADA under the same animation drivers for qualitative comparison.}

\begin{figure*}[tbh]
    \centering
    \includegraphics[width=0.95\linewidth]{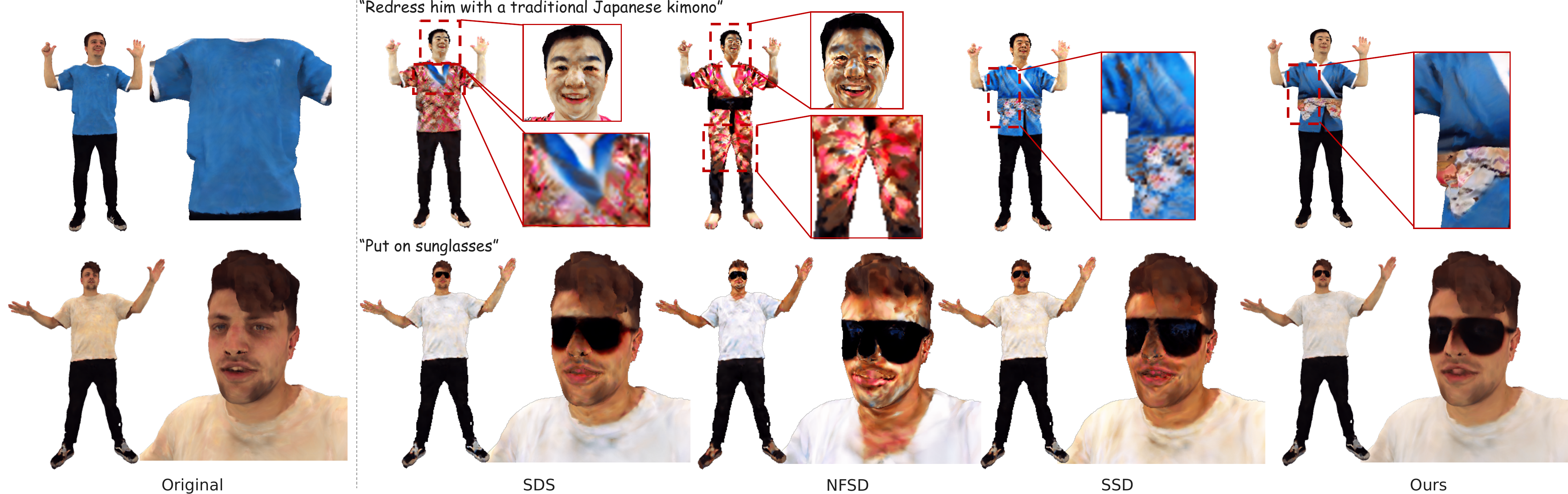}
    \caption{\small \textbf{Qualitative comparison with SDS, NFSD, and SSD}. Our method excels in both texture quality and adherence to the original avatars and editing instructions, whereas the others produce textures that are spotty, blurry, and over-saturated. }
    \label{fig:qual-comp-sds}
\end{figure*}

\begin{figure*}[tbh]
    \centering
    \includegraphics[width=\linewidth]{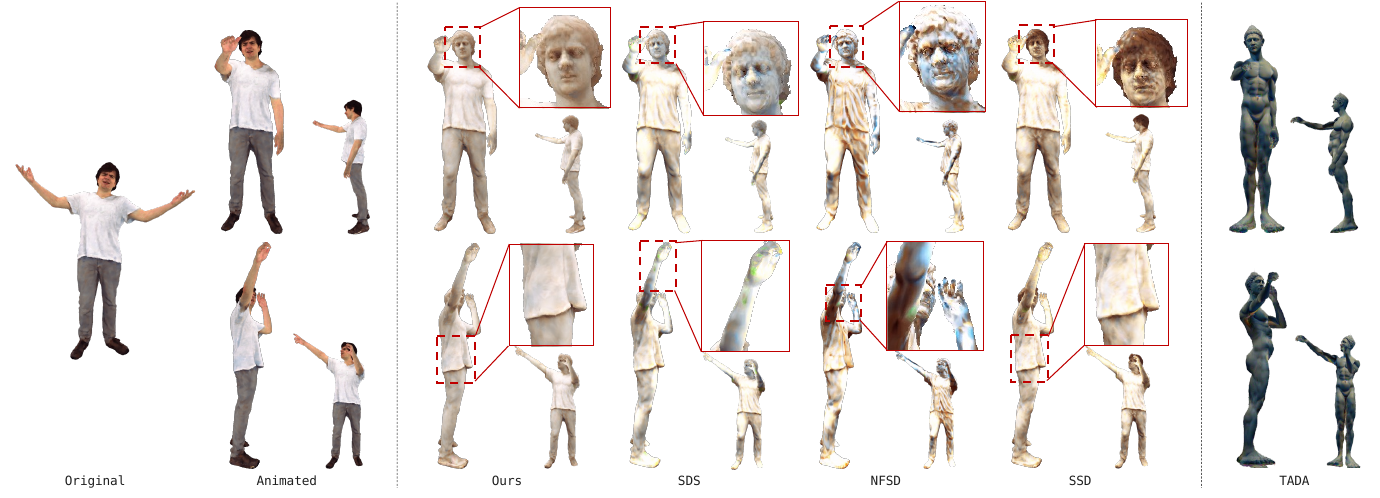}
    \caption{\rv{\small \textbf{Animation comparison} of edited or generated avatars.  We show the prompt ``A marble statue'' (full-body texture case). SDS, NFSD, and SSD are instantiated in the same framework~\cite{ho2023custom} as our method, making all edited avatars animatable; all results are driven by identical motions. These baselines exhibit spotty or over-saturated textures under animation. TADA~\cite{liao2024tada} produces animatable avatars from prompts but does not edit a given source avatar.}}
    \label{fig:animation-comparison}
\end{figure*}

\noindent\textbf{Implementation Details.}
\rv{We optimize for $1000$ steps and sample $50$ camera views per step at $400\times400$ resolution. Image conditioning follows IP2P~\cite{brooks_instructpix2pix_2023}: the source image is encoded by the Stable Diffusion VAE, and the resulting latent is concatenated with the noisy latent along the channel dimension after modifying the UNet’s first convolution to accept the additional channels. The pre-trained NeRF and initial human latent codes are identical to EditableHumans~\cite{ho2023custom}. Concretely, the NeRF uses two MLP decoders (SDF and RGB), each with $4$ linear layers and $128$ hidden units, and the latent is a $32$-D per-vertex codebook over $10{,}475$ mesh vertices. The smoothness loss weight is $w_1=300$. Gradient-aware viewpoint sampling is computed once at the first editing iteration (negligible overhead, $<0.1$\,s) and then fixed. With batch size $1$, the per-iteration time on an NVIDIA A40 is $25.2$\,s (total $\approx7$\,h for $1000$ steps) with peak memory $\approx13$\,GB.}

\subsection{Qualitative Experiment}\label{sec:qual}
We compare with IN2N and SDS-based methods using human subjects from CustomHumans~\cite{ho2023custom}. As shown in Fig.~\ref{fig:qual-comp-in2n} and Fig.~\ref{fig:qual-comp-sds}, our approach outperforms existing methods in producing high-quality textures that better follow editing instructions while retaining human identity. Unlike IN2N, our avatars remain fully animatable. %
For avatar generation baselines, Fig.~\ref{fig:qual-comp-avatarclip} shows that our method yields more photorealistic results compared to AvatarCLIP and TADA. 
\rv{Fig.~\ref{fig:animation-comparison} shows animated comparisons, where methods are evaluated under identical motions.}
More qualitative results are provided in Fig.~\ref{fig:qual_results}. %

\begin{figure}[!htbp]
    \centering
    \includegraphics[width=\linewidth]{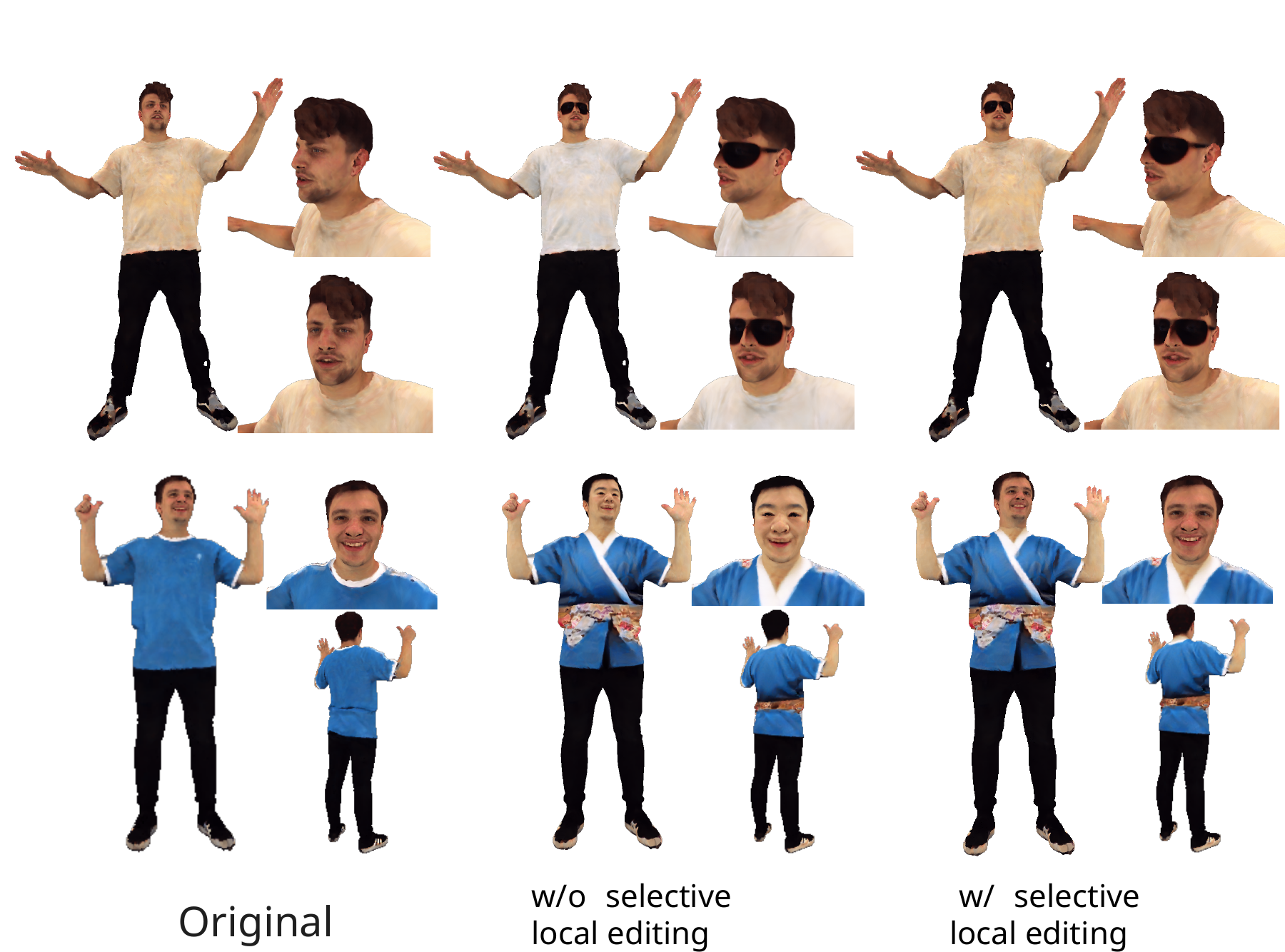}
    \caption{\small \rv{\textbf{Selective local editing}. Examples with head-only (``Put on a pair of sunglasses'') and clothes-only (``Redress him with a traditional Japanese kimono'') updates. This approach preserves the original clothing, improving upon the results from our primary pipeline.}}
    \label{fig:appen-local-edit}
\end{figure}

\noindent\rv{\textbf{Selective Local Editing.}
We also evaluate the alternative selective local editing. %
While the primary pipeline already achieves robust localized edits without affecting unselected areas, this option enhances precision when exact regions are specified.
As shown in Fig.~\ref{fig:appen-local-edit}, this method preserves the original clothing and identity cues, offering finer control over local edits compared to running the main pipeline alone, \eg the instruction ``Put on a pair of sunglasses,'' where the head region is edited and the rest of the body remains unchanged; for ``Redress him with a traditional Japanese kimono,'' only the clothes region is edited.}

\begin{figure}[!htbp]
    \centering
    \includegraphics[width=\linewidth]{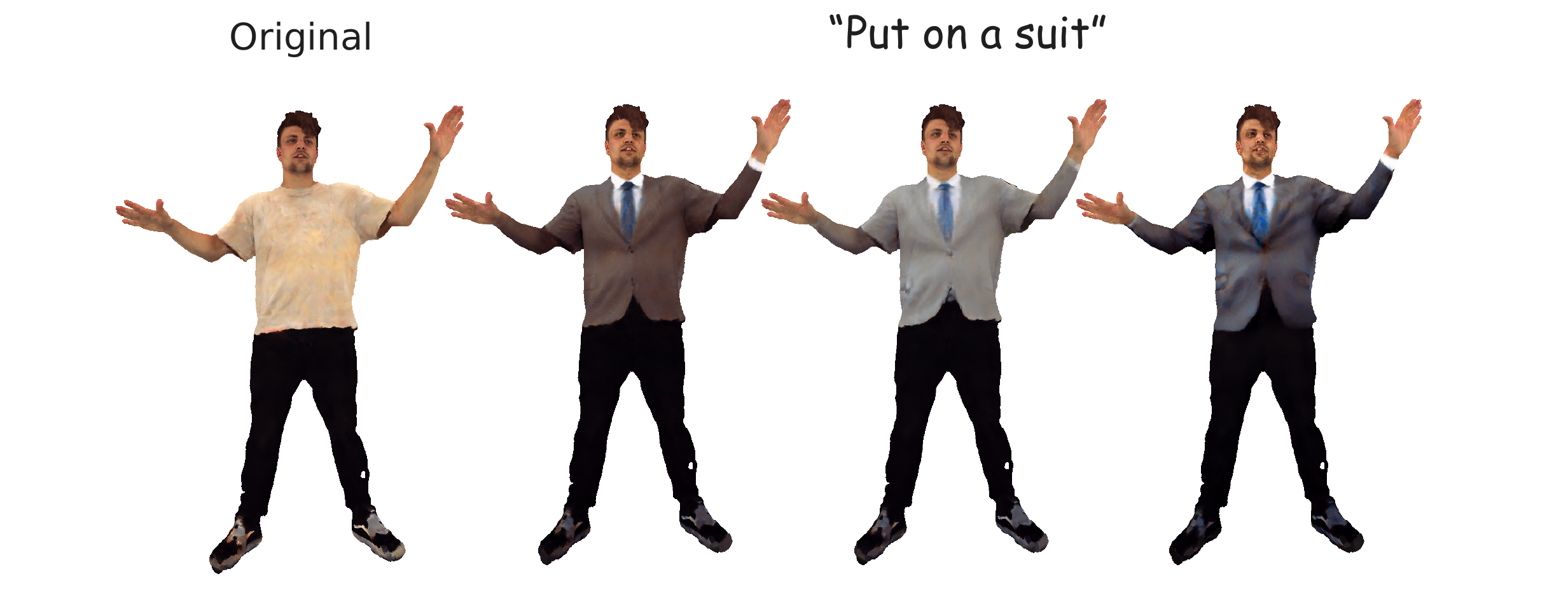}
    \caption{\small \textbf{Diversified results.}}
    \label{fig:appen-diversity}
\end{figure}

\noindent\textbf{Diversity.}
Most SDS-based methods exhibit limited diversity due to the inherent constraints of the distillation process~\cite{katzir2024nfsd}. While our work prioritizes editing quality, diversity can also be improved through a simple filtering strategy that amplifies edits with specific attributes, such as color (see Fig.~\ref{fig:appen-diversity}).

\subsection{Quantitative Experiment}\label{sec:quan}
\noindent\textbf{Metrics}. Following IN2N, we measure \textit{CLIP text-image directional similarit}y (CLIP-Direc$\uparrow$) %
for text alignment.
To evaluate structural and semantic fidelity to the original avatar, we also evaluate \textit{CLIP image similarity} (CLIP-Img$\uparrow$) between the rendered images of the edited and original avatars. %
Both metrics, in conjunction, balance maintaining consistency with the original image and achieving the intended editing outcomes. %
We also evaluate {LPIPS$\downarrow$}~\cite{zhang2018perceptual} {compared with the original images} for texture quality, and CLIP-Score$\uparrow$ between the result images and text prompts for generation coherence.

\noindent\textbf{Quantitative comparison with SOTA.}
We follow IN2N’s 10 full-body edits and add 3 localized edits (hair, eyes, mouth) for fine-grained evaluation.
As shown in Tab.~\ref{tab:quant_comp_sds}, %
we outperform IN2N and existing SDS-based methods %
across all metrics, achieving the best balance between text adherence, identity preservation, and visual quality.  
For avatar generation, Tab.~\ref{tab:quant_comp_avatarclip} shows our higher CLIP-Score, indicating better semantic alignment than AvatarCLIP and TADA.

\begin{table}[tbh]
\caption{\small \textbf{Quantitative comparison with text-based editing methods.}  SSD has slightly higher CLIP-Img at the cost of CLIP-Direc. Ours balances the best between adherence to instructions and preservation of original avatar features, while delivering superior texture quality.}
    \vspace{-0.8em}
    \label{tab:quant_comp_sds}
    \centering
    \scriptsize
    \begin{tabularx}{\linewidth}{>{\centering\arraybackslash}p{1.5cm}
                           *{5}{>{\centering\arraybackslash}X}}
        \toprule
 & Ours & IN2N & SDS & SSD & NFSD \\
 \midrule
CLIP-Direc$\uparrow$ & \textbf{0.160} & 0.117 & 0.144 & \textcolor{red}{0.141} & 0.149  \\
CLIP-Img$\uparrow$ & \textbf{0.863} & 0.686 & 0.852 & \textbf{0.868} & 0.783  \\
LPIPS$\downarrow$ & \textbf{0.047} & 0.216 & 0.070 & 0.057 & 0.109 \\
        \bottomrule
    \end{tabularx}
\end{table}
\begin{table}[tbh]
    \caption{\small \textbf{Quantitative comparison with avatar generation methods.} Ours achieves the highest CLIP-Score, showing superior semantic alignment.
    }
    \vspace{-0.8em}
    \label{tab:quant_comp_avatarclip}
    \centering
    \scriptsize
    \begin{tabularx}{\linewidth}{*{4}{>{\centering\arraybackslash}X}}%
        \toprule
         & Ours & AvatarCLIP & TADA \\
        \midrule
CLIP-Score$\uparrow$ & \textbf{0.231} & 0.223 & 0.230 \\
        \bottomrule
    \end{tabularx}
\end{table}

\begin{table}[tbh]
    \caption{\small \textbf{User study.} %
    Ours is significantly preferred across all three metrics. ``No Pref.'' indicates no preference. 
    }%
    \label{tab:user_study}
    \centering
    \scriptsize
    \setlength{\tabcolsep}{4pt}
    \begin{tabularx}{\linewidth}{l *{5}{>{\centering\arraybackslash}X}}%
        \toprule
              & {Ours} & NFSD & SDS & SSD & No Pref. \\
        \midrule
        Visual Quality & \textbf{57.82\%} & 10.26\% & 10.64\% & 19.42\% & 1.86\% \\
        Image Consistency& \textbf{58.14\%} & 7.50\% & 9.10\% & 22.63\% & 2.63\% \\
        Text Consistency& \textbf{53.59\%} & 15.96\% & 9.23\% & 19.23\% & 1.99\% \\
        \bottomrule
    \end{tabularx}
\end{table}%

\noindent\textbf{User Study.}
Editing quality is subjective. 
We conducted a {user study} %
on Mechanical Turk, 
where 315 participants provided 1560 responses on all editing comparisons, considering overall quality, instruction adherence, and fidelity to the original images.
As summarized in Tab.~\ref{tab:user_study}, our method is preferred across all metrics. 

\subsection{Ablation Studies}\label{sec:ablation} 
\noindent\rv{\textbf{Ablation setup.} Unless otherwise noted, all ablations are run with both text prompts and image conditioning (the source avatar). We vary one component at a time and keep all other settings fixed (same seed, prompt, and avatar).} %

\begin{figure}[!htbp]
    \centering
    \includegraphics[width=\linewidth]{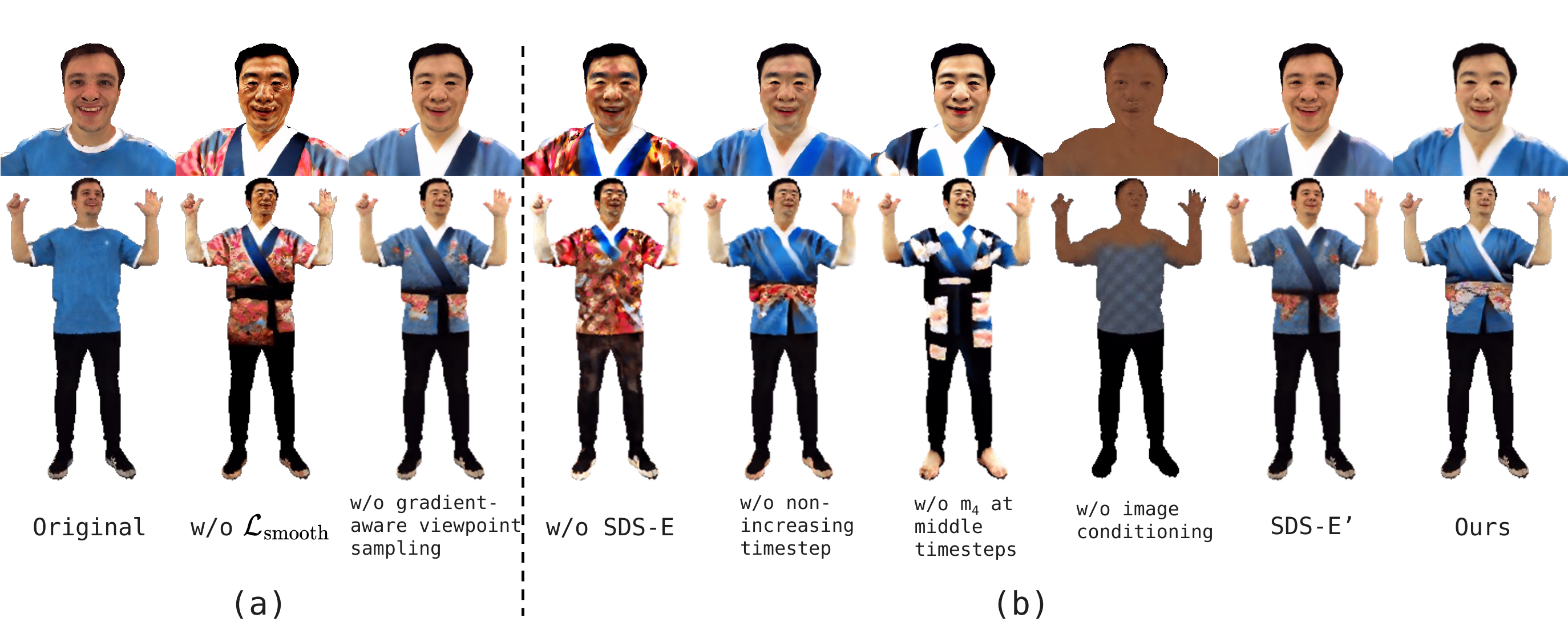}
    \vspace{-1em}
    \caption{\small \textbf{Ablation study on key components.}}
    \label{fig:ablation}
\end{figure}

\noindent \textbf{Smoothness regularizer \& Viewpoint sampling.} Fig.~\ref{fig:ablation} (a) %
shows how removing the regularizer leads to uneven textures and unrealistic spots, especially on the face. Omitting the {gradient-aware viewpoint sampling} leads to an undesired shift in the edited face due to an imprecise editing focus. Excluding it also increases the runtime 5-fold. %

\begin{figure}[!htbp]
    \centering
    \includegraphics[width=\linewidth]{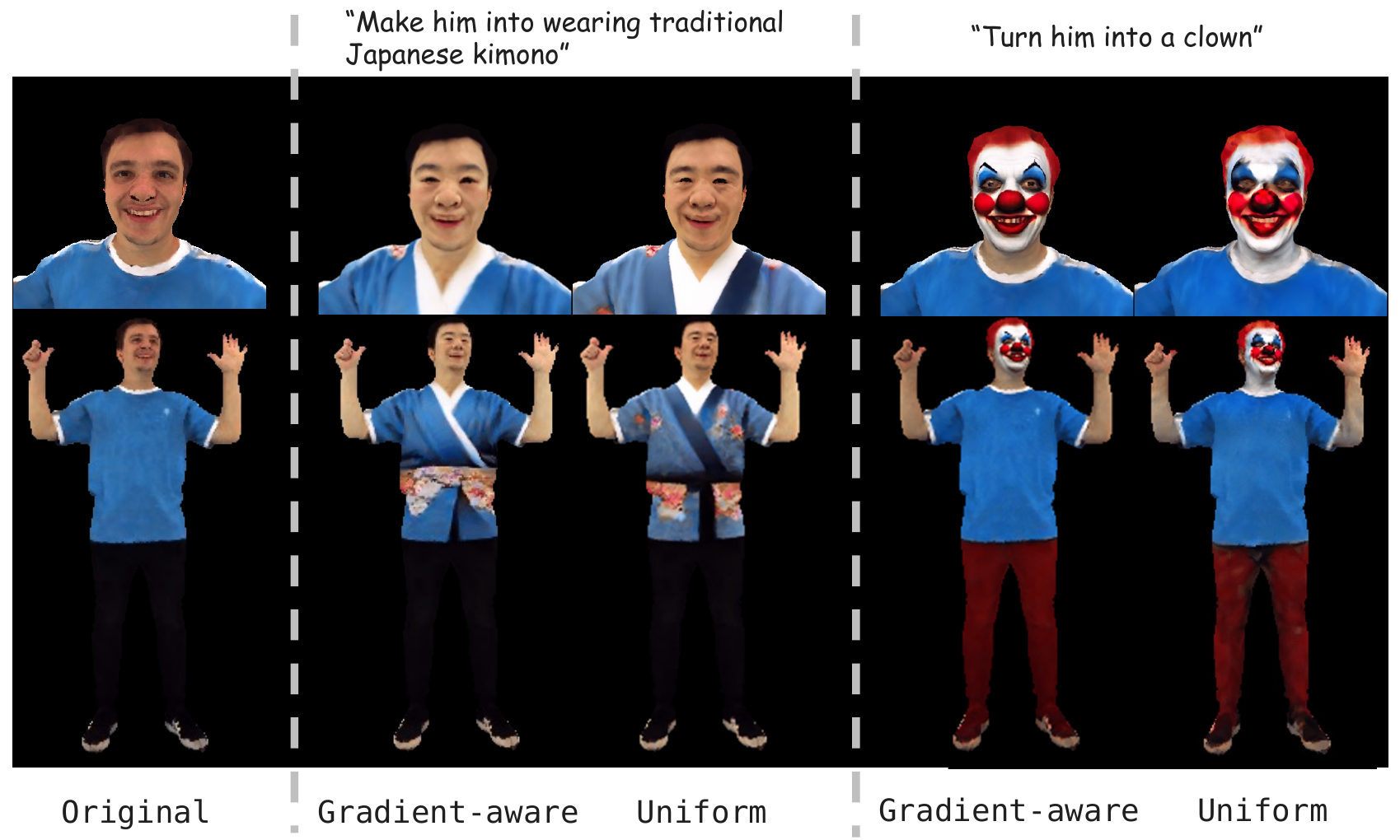}
    \caption{\small \textbf{Ablation study on gradient-aware viewpoint sampling.}  Our approach efficiently focuses editing on desired regions, whereas uniform sampling yields inaccurate and blurry results.}
    \label{fig:appen-viewsampl}
\end{figure}

\noindent \textbf{Gradient-Aware Viewpoint Sampling.}\label{appen:gradient-aware-view}
Figure~\ref{fig:appen-viewsampl} compares our gradient-aware sampling to a uniform baseline. %
Our method assigns region-specific weights $w_r$ (see Eq.~\ref{eq:view-weight}) and view counts $\mathcal C(r)$ (see Eq.~\ref{eq:view-count}), as detailed in Tab.~\ref{tab:appen-view-weights}. \rv{To validate the efficiency of the gradient-aware viewpoint sampling, we investigate its runtime. It is computed once at the first iteration, adding negligible overhead ($<$ 0.1 s) because it reuses gradients already computed in that iteration; the operation is dominated by simple reductions. Rather than the per-step cost, it reduces the overall iteration numbers until convergence, leading to $2\times$ fewer steps in overall runtime.}
Our sampling strategy successfully assigns weights to specific regions based on the editing instructions, ensuring more precise and efficient editing.

\begin{figure*}[!htbp]
    \centering
    \includegraphics[width=\linewidth]{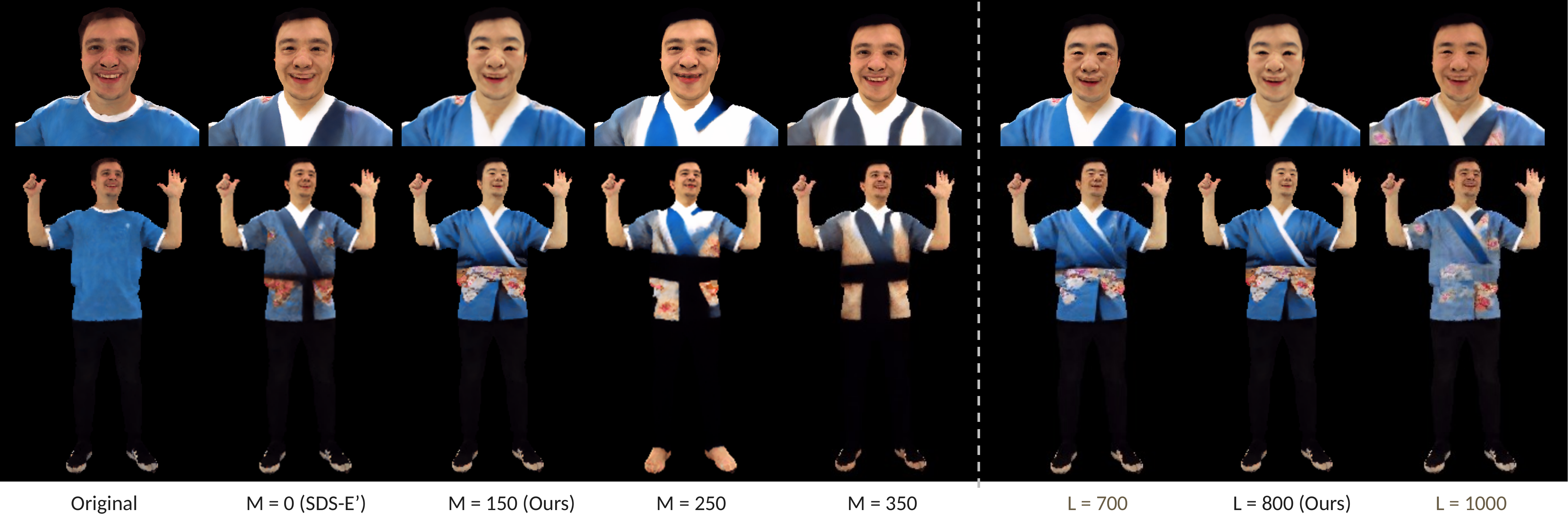}
    \caption{\small \rv{\textbf{Ablation study on timestep division thresholds.} \textbf{Left:} Varying $M$, the threshold between small and medium timesteps. Larger $M$ values lead to fewer details and possible editing failures. At $M=150$, results are most faithful to the editing instructions with high texture quality. When $M=0$ (equivalent to $\text{SDS-E}'$), results maintain similar quality but emphasize coherence with the original avatar. \textbf{Right:} Varying $L$, the threshold between medium and large timesteps. Larger $L$ values can destroy original features (e.g., clothing color), consistent with our analysis in Sec.~\ref{sec:analysis}. Setting $L=800$ gives the best results, while $L=700$ causes a slight drop in texture quality, noticeable in areas like the eyes.}
}
    \label{fig:appen-timestep-division}
\end{figure*}

\noindent\rv{\textbf{Timestep Division.}\label{appen:timestep-division}
We empirically determined the thresholds $M$ and $L$ for dividing timesteps into small, medium, and large stages (see Tab.~\ref{tab:sdse_components}). Fig.~\ref{fig:appen-timestep-division} illustrates the impact of these divisions on editing performance.}

\noindent \textbf{Other design choices.}  Fig.~\ref{fig:ablation} (b) explores various design choices. Without SDS-E (\ie using standard SDS) significantly damages the original clothing and facial features and produces saturation.
Omitting non-increasing timestep sampling adversely affects the convergence of clothing details, a consequence of intermediate traps detailed in Sec.~\ref{sec:interm_trap}.
An alternative approach that excludes term $m_4$ during middle timesteps leads to deviations from desired image guidance. \rv{Excluding image conditioning collapses to low-frequency color fill.}  
Comparing our default $\mathcal{L}_\text{SDS-E}$ with its alternative, $\mathcal{L}'_\text{SDS-E}$, the former achieves a balance between editing instructions and image adherence, while the latter preserves greater consistency with the original image, as observed in facial features. Therefore, we recommend a selective application of both loss functions, tailored to the specific editing contexts.

\begin{table}[!htbp]
\scriptsize
\caption{\small \textbf{Region weights $w_r$ and view counts $\mathcal{C}(r)$ from our gradient-aware sampling.} The total number of views is $|V|=50\,000$. For the ``kimono'' instruction, body regions receive higher weights and more views to match its focus on clothing; for ``clown,'' more views are assigned to the head for intensive editing.}
\label{tab:appen-view-weights}
\centering
\resizebox{\columnwidth}{!}{%
\begin{tabular}{c|c|ccccc}
\toprule
\multirow{2}{*}{Instruction} & \multirow{2}{*}{Metric} & \multicolumn{5}{c}{Region} \\ \cline{3-7} 
                             &                         & Face & Back Head & Front Body & Back Body & Arms \\ \hline
\multirow{2}{*}{``kimono''}   & Weight      & 0.04 & 0.08     & 0.47       & 0.26      & 0.15 \\ 
                             & View Count & 2000 & 4000     & 23500      & 13000     & 7500 \\ \hline
\multirow{2}{*}{``clown''}    & Weight      & 0.07 & 0.20     & 0.30       & 0.31      & 0.12 \\ 
                             & View Count & 3500 & 10000    & 15000      & 15500     & 6000 \\ \bottomrule
\end{tabular}%
}
\end{table}

\subsection{Applications}\label{sec:applications}
\begin{figure*}[!htbp]
    \centering
    \includegraphics[width=\linewidth]{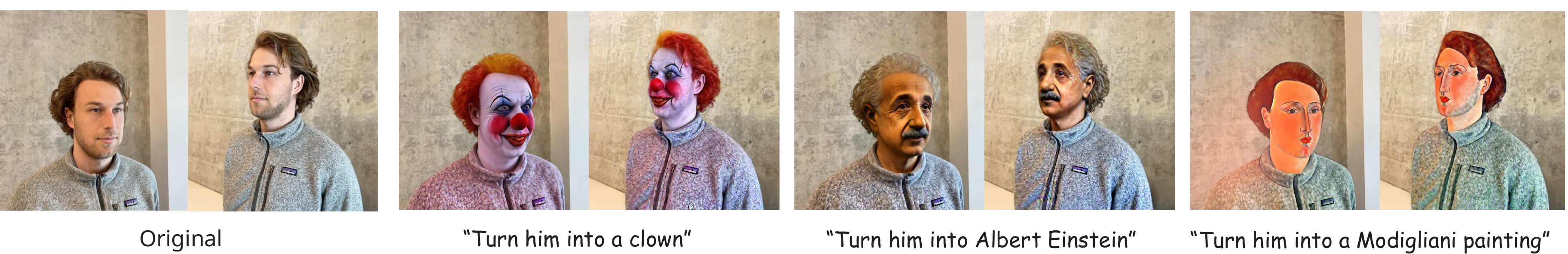}
    \caption{\small Applying SDS-E on the GaussianEditor~\cite{Chen2023GaussianEditor} pipeline, which enables simultaneous texture and geometry editing. %
    }
    \label{fig:appen-sdse-gaussianeditor}
\end{figure*}

InstructHumans and SDS-E enable various applications. Leveraging the explicit controllability of the human representation~\cite{ho2023custom}, InstructHumans allows \textit{animation} of edited avatars by adjusting SMPL-X mesh's pose parameters (see Fig.~\ref{fig:appen-animation}). %
Furthermore, SDS-E can be applied to broader pipelines, such as \textit{editing texture and geometry together} on a 3D Gaussian splatting pipeline~\cite{Chen2023GaussianEditor} (see Fig.~\ref{fig:appen-sdse-gaussianeditor}).

\begin{figure}[!htbp]
    \vspace{-1.5em}
    \includegraphics[width=\linewidth]{svg-inkscape/appen-animation_svg-raw.pdf}
    \vspace{-2em}
    \caption{\small \textbf{Animated edited avatars}. The animations demonstrate the seamless integration of editing while preserving natural motion.}
    \label{fig:appen-animation}
\end{figure}

\subsection{Limitations.}
\begin{figure}[!htbp]
    \centering
    \includegraphics[width=\linewidth]{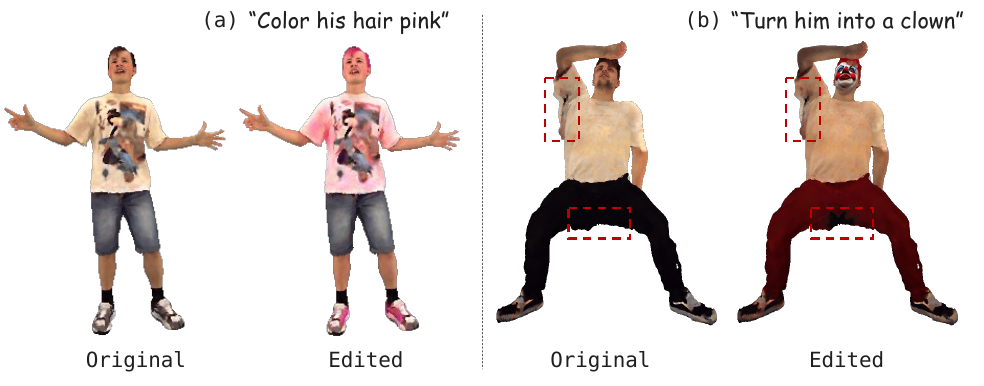}
    \caption{\small \rv{\textbf{Failure cases.} (a) Global color leakage when IP2P is prompted with explicit colors. (b) Joint artifacts under extreme poses due to representation limits.}
    }
    \label{fig:failure-case}
\end{figure}

\rv{Fig.~\ref{fig:failure-case} depicts typical failure cases.
Our framework employs IP2P~\cite{brooks_instructpix2pix_2023} to guide texture edits and thus inherits its behavior. A commonly observed issue with IP2P is that, when prompts specify colors, it tends to apply a global color cast that spills beyond the intended region, as illustrated in Fig.~\ref{fig:failure-case}(a). Our selective local editing mitigates this to some extent, but a more thorough remedy is to adopt stronger 2D editing models as they become available.
In addition, our framework builds upon the hybrid human representation of~\cite{ho2023custom}, which may produce artifacts at joint areas under extreme poses, as seen in Fig.~\ref{fig:failure-case}(b). As noted in~\cite{ho2023custom}, increasing mesh resolution and training on larger datasets are potential remedies. Our framework can directly benefit from such improvements.}

\section{Conclusion}
This work presents a method for 3D human texture editing guided by textual instructions, achieving a balance between intuitive editing capabilities and animation flexibility. By analyzing and adapting SDS, we propose SDS for Editing (SDS-E), to distill faithful and high-fidelity editing guidance from the 2D diffusion model. Enhancements including Laplacian latent smoothness and gradient-aware viewpoint sampling further augment the efficiency and effectiveness of our editing pipeline. Experiments affirm our method's superior editing performance relative to existing text-based 3D editing approaches.%

\bibliographystyle{IEEEtran}
\bibliography{ref}

\vfill

\end{document}